\documentclass{article}

\usepackage{arxiv}

\usepackage{url}            
\usepackage[pdftex,hyperfootnotes=true,pdfpagelabels,linktocpage,
colorlinks=true,
allcolors=blue,
]{hyperref}  

\usepackage[utf8]{inputenc} 
\usepackage[T1]{fontenc}    
\usepackage{booktabs}       
\usepackage{amsfonts}       
\usepackage{nicefrac}       
\usepackage{microtype}      
\usepackage{lipsum}		
\usepackage{graphicx}
\usepackage{doi}
\usepackage{algorithm}

\input{some_cmds}

\renewcommand{\shorttitle}{Published as a conference paper at LOD 2020\hfill}

\renewcommand{\headeright}{}
\renewcommand{\undertitle}{}

\begin{document}
\title{State Representation Learning from Demonstration}
%
%
\author{Astrid Merckling, 
Alexandre Coninx, 
Loic Cressot, 
Stéphane Doncieux, Nicolas Perrin-Gilbert\\ 
\\
ISIR, Institut des Systèmes Intelligents et de Robotique\\Sorbonne University, CNRS\\Paris, France\\
%
\texttt{astrid.merckling@sorbonne-universite.fr}
}
\date{}

\maketitle              
\begin{abstract}

Robots could learn their own state and world representation from perception and experience without supervision. This desirable goal is the main focus of our field of interest, state representation learning (SRL). Indeed, a compact representation of such a state is beneficial to help robots grasp onto their environment for interacting. The properties of this representation have a strong impact on the adaptive capability of the agent.
In this article we present an approach based on imitation learning. The idea is to train several policies that share the same representation to reproduce various demonstrations. To do so, we use a multi-head neural network with a shared state representation feeding a task-specific agent.
If the demonstrations are diverse, the trained representation will eventually contain the information necessary for all tasks, while discarding irrelevant information. As such, it will potentially become a compact state representation useful for new tasks. We call this approach SRLfD (State Representation Learning from Demonstration).
Our experiments confirm that when a controller takes SRLfD-based representations as input, it can achieve better performance than with other representation strategies and promote more efficient reinforcement learning (RL) than with an end-to-end RL strategy.

\end{abstract}

\section{Introduction}
\label{sec:intro}

Recent reinforcement learning (RL) achievements might be attributed to a combination of (i) a dramatic increase of computational power, (ii) the remarkable rise of deep neural networks in many machine learning fields including robotics, which take advantage of the simple idea that training with quantity and diversity helps. The core idea of this work consists of leveraging task-agnostic knowledge learned from several task-specific agents performing various instances of a task.

Learning is supposed to provide animals and robots with the ability to adapt to their environment. RL algorithms define a theoretical framework that is efficient on robots~\citep{kober2013surveyRL} and can explain observed animal behaviors~\citep{schultz1997neural}. These algorithms build policies that associate an action to a state to maximize a reward. The state determines what an agent knows about itself and its environment. A large state space -- raw sensor values, for instance -- may contain the relevant information but would require a too large exploration to build an efficient policy. Well-thought feature engineering can often solve this issue and make the difference between the failure or success of a learning process.
In their review of representation learning, \citet{bengio2013RepLreview} formulate the hypothesis that the most relevant pieces of information contained in the data can be more or less entangled and hidden in different representations. If a representation is adequate, functions that map inputs to desired outputs are somewhat less complex and thus easier to construct via learning. However, a frequent issue is that these adequate representations may be task-specific and difficult to design, and this is true in particular when the raw data consists of images, \ie 2D arrays of pixels. One of the objectives of deep learning methods is to automatize feature engineering to make learning algorithms effective even on raw data. By composing multiple nonlinear transformations, the neural networks on which these methods rely are capable of progressively creating more abstract and useful representations of the data in their successive layers.


The intuition behind our work is that many tasks operated in the same environment share some common knowledge about that environment. This is why learning all these tasks with a shared representation at the same time is beneficial.
The literature in imitation learning \citep{pastor2009learning,kober2012reinforcement} has shown that demonstrations can be very valuable to learn new policies. To the best of our knowledge, no previous work has focused on constructing reusable state representations from raw inputs solely from demonstrations, therefore, here we investigate the potential of this approach for SRL.

\begin{figure}[t!]
\centering
\subcaptionbox{
Preliminary\\ phase\label{fig:SRLfD_a}}{\raisebox{1.1cm}{%
\includegraphics[width=0.18\linewidth]{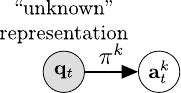}}}
\hfill
\subcaptionbox{
Pretraining\\ phase\label{fig:SRLfD_b}}{\includegraphics[width=0.3\linewidth]{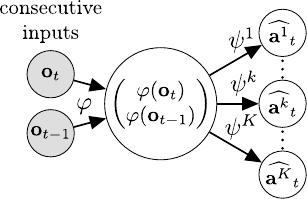}}
\hfill
\subcaptionbox{
Transfer learning\\  \hspace*{0.6cm} phase\label{fig:SRLfD_c}}{\raisebox{0.6cm}{%
\includegraphics[width=0.3\linewidth]{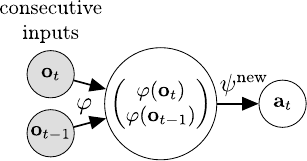}}}
\caption{\label{fig:SRLfD}
SRLfD (State Representation Learning from Demonstration) consists of three phases.
\textbf{\subref{fig:SRLfD_a}} Preliminary phase: for $K$ different tasks, we assume to have access to oracle policies ($\pi^k$) that solve each task, and compute their outputs with an ``unknown'' state representation.
\textbf{\subref{fig:SRLfD_b}} Pretraining phase: learning of one shared representation function $\varphi$ with imitation learning of $K$ specific heads $\psi^k$ by observing $\pi^k$ from high-dimensional observations.
Each head $\psi^k$ defines a sub-network that contains the parameters $\*\theta_\varphi$ of $\varphi$ and the parameters $\*\theta_{\psi^k}$ of $\psi^k$. The set of all network's parameters is
$\*\theta =\{\*\theta_\varphi,\*\theta_{\psi^1},\dots,\*\theta_{\psi^K}\}$.
\textbf{\subref{fig:SRLfD_c}} Transfer learning phase: the pretrained network $\varphi$ provides representations to learn an unseen decision making task $\psi^{\text{new}}$.}
\end{figure}

In this paper, we are interested in solving continuous control tasks via RL or supervised learning, using state estimates as inputs, without having access to any other sensor, which means in particular that the robot configuration, which we will call \emph{ground truth representation}, is unknown.
We assume that at all times the consecutive high-dimensional observations ($\:o_{t-1},\, \:o_t$) contain enough information to know the ground truth state $\:q_t$ and that the controller/predictor only needs to rely on this representation to choose actions.
Intuitively, $\:q_t$ could probably be a much better input for a RL algorithm than the raw images, but without prior knowledge, it is not easy to get $\:q_t$ from ($\:o_{t-1}$,$\:o_t$). In robotics, SRL~\citep{lesort2018surveySRL} aims at constructing a mapping from high-dimensional observations to lower-dimensional representations which, similarly to $\:q_t$, can be advantageously used instead of ($\:o_{t-1}$,$\:o_t$) to form the inputs of a policy.

Our proposed experimental setup consists in three different phases:
\begin{enumerate}
\item Preliminary phase (\fig{fig:SRLfD}\subref{fig:SRLfD_a}): we have $K$ controllers called oracle policies $\pi^k$, each solving a different task. For example, we could define them in laboratory conditions with better sensors (e.g. motion capture), the goal being to reproduce them with a different perception (e.g. images) where in this setting, building a representation extracted from the raw inputs makes sense. For the sake of the experiments, we used almost fake tasks.
\item Pretraining phase (\fig{fig:SRLfD}\subref{fig:SRLfD_b}): we derive a state representation that can be relied on to reproduce any of these oracle policies. We do so via imitation learning on a multi-head neural network consisting of a first part that outputs a common state representation $\:s_{t}\triangleq\big[\varphi(\:o_{t-1}),\varphi(\:o_t)\big]$ used as input to $K$ heads $\psi^k$ trained to predict actions $\:a_t^k$ executed by the oracle policies $\pi^k$ from the previous phase.
\item Transfer learning phase (\fig{fig:SRLfD}\subref{fig:SRLfD_c}): we use the previously trained representation $\:s_{t}$ as input to a new learning process $\psi^{\text{new}}$ in the same environment.
\end{enumerate}
This method, which we call SRLfD (State Representation Learning from Demonstration), is presented in more detail in \mysec{sec:SRLfD_method}, after an overview of the existing related work in \mysec{sec:rw}.
We show that using SRLfD learned representations instead of raw images can significantly accelerate RL (using the popular SAC algorithm \citep{haarnoja2018soft}). When the state representation is chosen to be low-dimensional, the speed up brought by our method is greater than the one resulting from state representations obtained with deep autoencoders, or with principal component analysis (PCA).

\section{Related Work}
\label{sec:rw}
SRL for control is the idea of extracting from the sensory stream the information that is relevant to control the robot and its environment and representing it in a way that is suited to drive robot actions. It has been subject to a lot of recent attention~\citep{lesort2018surveySRL}. It was proposed as a way to overcome the curse of dimensionality, to speed up and improve RL, to achieve transfer learning, to ignore distractors, and to make artificial autonomous agents more transparent and explainable~\citep{de2018integratingSRL,lesort2018surveySRL}.

Since the curse of dimensionality is a major concern, many state representation techniques are based on dimension reduction~\citep{kober2013surveyRL,bengio2013RepLreview} and traditional unsupervised learning techniques such as principal component analysis (PCA)~\citep{curran2015using} or its nonlinear version, the autoencoder~\citep{hinton2006reducing}. Those techniques allow to compress the observation in a compact latent space, from which it can be reconstructed with minimal error. Further developments led to variational autoencoders (VAE)~\citep{kingma2014reparam} and then their extension $\beta$-VAE~\citep{higgins2017beta}, which are powerful generative models able to learn a disentangled representation of the observation data. However, the goal of those methods is to model the observation data; they do not take actions into account, and the representation they learn is optimized to minimize a reconstruction loss, not to extract the most relevant information for control. In particular, their behavior is independent of physical properties, or the temporal structure of transitions, and they cannot discriminate distractors.

To overcome this limitation, a common approach to state representation is to couple an autoencoder to a forward model predicting the future state \citep{watter2015E2C}. A different approach to state representation is to forego observation reconstruction and to learn a representation satisfying some physics-based priors like temporal coherence, causality, and repeatability~\citep{jonschkowski2015priors} or controllability \citep{jonschkowski2017pves}. Those methods have been shown to learn representations able to speed up RL, but this improvement is contingent on the careful choice and weighting of the priors suited to the task and environment.

Learning state representations from demonstrations of multiple policies solving different tasks instances, as we propose, has some similarities with multi-task and transfer learning~\citep{taylor2009transferRL}. Multi-task learning aims to learn several similar but distinct tasks simultaneously to accelerate the training or improve the generalization performance of the learned policies, while transfer learning strives to exploit the knowledge of how to solve a given task to then improve the learning of a second task. Not all multi-task and transfer learning works rely on explicitly building a common representation, but some do, either by using a shared representation during multiple task learning~\citep{pinto2017learning} or by distilling a generic representation from task-specific features~\citep{rusu2015policyDistillation}. The common representation can then be used to learn new tasks. However, all these techniques rely on the end-to-end RL approach, which is less sample-efficient than the self-supervised learning approach followed by SRLfD.

Learning state representations from demonstrations of multiple policies solving different tasks instances, as we propose, has some similarities with multi-task and transfer learning~\citep{taylor2009transferRL}. Multi-task learning aims to learn several similar but distinct tasks simultaneously to accelerate the training or improve the generalization performance of the learned policies, while transfer learning strives to exploit the knowledge of how to solve a given task to then improve the learning of a second task. Not all multi-task and transfer learning works rely on explicitly building a common representation, but some do, either by using a shared representation during multiple task learning~\citep{pinto2017learning} or by distilling a generic representation from task-specific features~\citep{rusu2015policyDistillation}. The common representation can then be used to learn new tasks. However, all these techniques rely on the end-to-end RL approach, which is less sample-efficient than the self-supervised learning approach followed by SRLfD.

In another perspective, the learning from demonstration literature typically focuses on learning from a few examples and generalizing from those demonstrations, for example by learning a parameterized policy using control-theoretic methods~\citep{pastor2009learning} or RL-based approaches~\citep{kober2012reinforcement}. Although those methods typically assume prior knowledge of a compact representation of the robot and environment, some of them directly learn and generalize from visual input~\citep{finn2017one} and do learn a state representation. However, the goal is not to reuse that representation to learn new skills but to produce end-to-end visuomotor policies generalizing the demonstrated behaviors in a given task space.
Several works have also proposed using demonstrations to improve regular deep RL techniques~\citep{vevcerik2017DDPGfD,nair2018overcoming},
but the goal is mostly to improve exploration in environments with sparse rewards. Those works do not directly address the problem of state representation learning.

\vfill

\section{State Representation Learning from Demonstration}
\label{sec:SRLfD_method}

\subsection{Demonstrations}
\label{sec:SRLfD_demonstrations}
Let us clarify the hierarchy of the objects that we manipulate and introduce our notations. This work focuses on simultaneously learning $K$ different tasks\footnote{Roughly, different tasks refer to goals of different natures, while different instances of a task refer to a difference of parameters in the task.
For example, reaching various locations with a robotic arm is considered as different instances of the same reaching task.} sharing a common state representation function $\varphi$ and with $K$ task-specific heads for decision $(\psi^1,\psi^2,\dots,\psi^K)$ (see \fig{fig:SRLfD}\subref{fig:SRLfD_b}). For each $k$-task, the algorithm has seen demonstrations in a form of paths $P_1^k, P_2^k, \dots, P_P^k$ from an initial random position to the same goal corresponding to the $k$-task generated by running the oracle policy $\pi^k$ obtained in the preliminary phase (see \fig{fig:SRLfD}\subref{fig:SRLfD_a}).
Specifically, during a path $P_p^k$, an agent is shown a demonstration (or data point) of $(\:o_{t-1}^{k,p}, \:o_t^{k,p}, \:a_t^{k,p})$ from which it can build its own world-specific representation. Here, $\:o_{t-1}^{k,p}$ and $\:o_t^{k,p}$ are consecutive high-dimensional observations (\aka measurements), and $\:a_t^{k,p}$ is a real-valued vector corresponding to the action executed right after the observation $\:o_t^{k,p}$ was generated.

\subsection{Imitation Learning from Demonstration}
\label{sec:SRLfD_training}

Following the architecture described in \fig{fig:SRLfD}\subref{fig:SRLfD_b}, we use a state representation neural network $\varphi$ that maps high-dimensional observations $\:o_t^{k,p}$ to a smaller real-valued vector $\varphi(\:o_t^{k,p})$.
This network $\varphi$ is applied to consecutive observations ($\:o_{t-1}^{k,p}$, $\:o_t^{k,p}$) to form the state representation $\:s_{t}^{k,p}$, as follows:
\begin{equation}\label{eq:concatenated_state}
\:s_{t}^{k,p} = \big[\varphi(\:o_{t-1}^{k,p}),\varphi(\:o_t^{k,p})\big]
\end{equation}
This state representation $\:s_{t}^{k,p}$ is sent to the $\psi^k$ network, where $\psi^k$ is one of the $K$ independent heads of our neural network architecture.
$\psi^1$, $\psi^2$, \dots, $\psi^K$ are head networks with similar structure but different parameters, each one corresponding to a $k$-task. Each head has continuous outputs with the same number of dimensions as the action space of the robot.
We denote by $\psi^k(\:s_{t}^{k,p})$ the output of the $k$-th head of the network on the input $(\:o_{t-1}^{k,p}, \:o_t^{k,p})$.
We train the global network to imitate all the oracle policies via supervised learning.
Specifically, our goal is to minimize the quantities: $\Vert \psi^k(\:s_{t}^{k,p}) - \:a_t^{k,p} \Vert_2^2$
that measure how well the oracle policies are imitated.
The optimization problem we want to solve is thus the minimization of the following objective function:
\begin{equation}\label{eq:srlfd_objective}
\mathcal{L}(\*\theta) = \frac{1}{PT}\sum_{p=1}^P\sum_{t=1}^T \Vert
\psi^k(\:s_{t}^{k,p}) - \:a_t^{k,p} \Vert_2^2
\end{equation}
for $k\in \llbracket 1,K \rrbracket$, and where $\*\theta =\{\*\theta_\varphi,\*\theta_{\psi^1},\dots,\*\theta_{\psi^K}\}$ as explained in \fig{fig:SRLfD}\subref{fig:SRLfD_b}.
We give an equal importance to all oracle policies by uniformly sampling $k \in\nobreak\llbracket 1,K \rrbracket$, and performing a training step on $\mathcal{L}(\*\theta)$ to adjust $\*\theta$. \algo{algo:SRLfD} describes this procedure.

\begin{algorithm}[t!]
\caption{SRLfD algorithm\label{algo:SRLfD}}
\begin{algorithmic}[1]
\STATE \textbf{Input:} A set of instances of tasks $T^k$, $k\in\nobreak\llbracket 1,K \rrbracket$, and for each of them a set of paths $P_p^k$, $p\in\nobreak\llbracket 1,P \rrbracket$ of maximum length $T$.
\STATE \textbf{Initialization:} A randomly initialized neural network following the architecture described in \fig{fig:SRLfD}\subref{fig:SRLfD_b} with parameters $\*\theta =\{\*\theta_\varphi,\*\theta_{\psi^1},\dots,\*\theta_{\psi^K}\}$.
\WHILE{$\*\theta$ has not converged}
\STATE Pick uniformly a $k$-task
\STATE Predict current state representations with \eq{eq:concatenated_state}:
\[\:s_{t}^{k,p}=
\big[\varphi(\:o_{t-1}^{k,p}),\varphi(\:o_t^{k,p})\big]\]
\STATE Compute $\mathcal{L}(\*\theta)$ with \eq{eq:srlfd_objective}:
\[\mathcal{L}\gets \frac{1}{PT}\sum_{p=1}^P\sum_{t=1}^T \Vert \psi^k(\:s_{t}^{k,p}) - \:a_t^{k,p} \Vert_2^2 \]
\STATE Perform a training step on $\mathcal{L}(\*\theta)$ w.r.t. $\*\theta$
\ENDWHILE
\end{algorithmic}
\end{algorithm}

The network of SRLfD is trained to reproduce the demonstrations, but without direct access to the ground truth representation of the robot. Each imitation can only be successful if the required information about the robot configuration is extracted by the state representation $\big[\varphi(\:o_{t-1}^{k,p}),\varphi(\:o_t^{k,p})\big]$. However, a single task may not require the knowledge of the full robot state. Hence, we cannot be sure that reproducing only one instance of a task would yield a good state representation. By learning a common representation for various instances of tasks, we increase the probability that the learned representation is general and complete. It can then be used as a convenient input for new learning tasks, especially for a RL system.

\section{Goal Reaching}

In this section, we study a transfer learning phase (see \fig{fig:SRLfD}\subref{fig:SRLfD_c}) corresponding to a RL optimization problem to solve a torque-controlled reaching task with image observations.
This is a challenging problem despite the simplicity of the task. Indeed, when high-dimensional observations are mapped to a lower-dimensional space before feeding a RL system, a lot of information is compacted and valuable information for control may be lost.
The purpose is to verify that state representations learned with SRLfD are useful representations for RL algorithms.
In this work, we only conduct experimental validations, but a fundamental question that we will not answer is at stake: what constitutes a good representation for state-of-the-art deep RL algorithms? Should it be as compact and as disentangled as possible, or on the contrary, can redundancy of information or correlations be useful in the context of deep RL? A definitive answer seems beyond the current mathematical understanding of deep RL.

\begin{figure}[t!]
\centerline{\includegraphics[width=0.9\columnwidth]{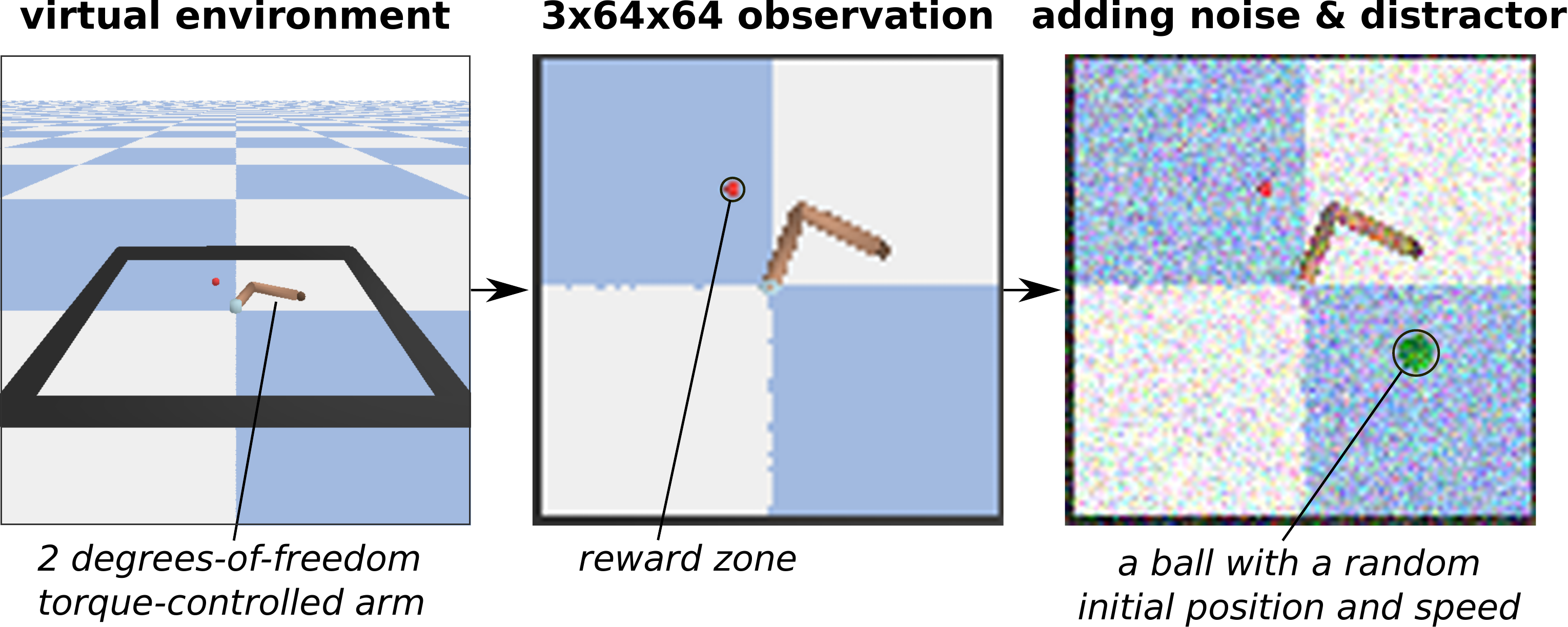}}
\caption{\label{fig:reacher}
The \emph{Reacher} environment, with a reward of $1$ when the end-effector reaches a position close to the goal, and $0$ otherwise. For more challenging inputs (on the right), we add Gaussian noise with zero mean and standard deviation 10, and a ball distractor is added in the environment with random initial position and velocity.}
\end{figure}

We consider a simulated 2D robotic arm with 2 torque-controlled joints, as shown in \fig{fig:reacher}. We use the environment \emph{Reacher} adapted from the OpenAI Gym~\citep{brockman2016openai} benchmark to PyBullet~\citep{coumans2018pybullet}. An instance of this continuous control task is parameterized by the position of a goal that the end-effector of the robot must reach within some margin of error (and in limited time). We use as raw inputs RGB images of 64$\times$64 pixels.
As the heart of our work concerns state estimation, we have focused on making perception challenging, by adding in some cases randomly moving distractor and Gaussian noise, as shown in \fig{fig:reacher}. We believe that the complexity of the control part (\ie the complexity of the tasks) is less important to validate our method, as it depends more on the performance of the RL algorithm. To solve even just the simple reaching task, the configuration of the robot arm is required and needs to be extracted from images for the RL algorithm to converge. Indeed our results show that this is the case when SRLfD learned representations are used as inputs of SAC~\citep{haarnoja2018soft}.

\subsection{Experimental Setup}
\label{sec:exp}

\paragraph{Baseline Methods}
We compare state representations obtained with our method (SRLfD) to five other representation strategies:
\begin{itemize}
\item Ground truth: as mentioned in Section~\ref{sec:exp}, what we call \emph{ground truth} representation of the robot configuration is represented as a vector of size four: the two torques angles and velocities.
\item Principal Component Analysis (PCA)~\citep{jolliffe2011principal}:
we perform PCA on the demonstration data, and the $8$ or $24$ most significant dimensions are kept, thus reducing observations to a compact vector that accounts for a large part of the input variability.
\item Autoencoder-based representation~\citep{hinton2006reducing}:
$\varphi$ is replaced by an encoder learned with an autoencoder.
The latent space representation of the autoencoder (of size $8$ or $24$) is trained with the same demonstrations (but ignoring the actions) as in the SRLfD training.
\item Random network representation~\citep{gaier2019weight}:
we use the same neural network structure for $\varphi$ as with SRLfD, but instead of training its parameters, they are simply fixed to random values sampled from a Gaussian distribution of zero mean and standard deviation $0.02$.
\item Raw pixels: the policy network is modified to receive directly $(\:o_{t-1},\:o_t)$ in input, with the same dimensionality reduction after $\varphi$ as other methods, but all of its parameters are trained simultaneously in the manner of end-to-end RL.
\end{itemize}
The representations obtained with these methods use the same demonstration data as SRLfD method, and share the same neural network structure for $\varphi$ or replace it (with ground truth and PCA) in the architecture of \fig{fig:SRLfD}\subref{fig:SRLfD_c} whose output size is $8$ or $24$\footnote{The number of $24$ dimensions has been selected empirically (not very large but leading to good RL results).}.


\paragraph{Generating Demonstrations}
For simplicity, the preliminary phase of training $K$ oracle policies $\pi^k$ (see \fig{fig:SRLfD}\subref{fig:SRLfD_a}) is done by running the SAC~\citep{haarnoja2018soft} RL algorithm. Here, the ``unknown'' representations used as inputs are the ground truth representations. SAC also exploits the cartesian cordinates of the goal position\footnote{The purpose of our method is to generate state representations
from (possibly noisy) inputs that are hard to exploit (such as raw images), so only the preliminary phase has access to the ``unknown'' representation.}.
It returns a parameterized policy capable of producing reaching trajectories to any goal position.

For the pretraining phase of SRLfD (see \fig{fig:SRLfD}\subref{fig:SRLfD_b}), the previously learned parameterized policy generates $K=16$ oracle policies $\pi^k$ (which represent different instances of the reaching task), with each of them $238$ paths for training and $60$ paths for validation of maximal length $T=50$, computed from various initial positions.
We then simultaneously train all the heads for computational efficiency. Specifically, for each optimization iteration, we uniformly sample for every head $\psi^k$ a mini-batch of 64 demonstrations from the $P$ paths corresponding to the $k$-task.

\paragraph{Implementation Details}
For SRLfD network architecture (adapted from the one used in~\citep{mnih2013playing}), $\varphi$ (see \fig{fig:SRLfD}) sends its $3\times 64\times 64$ input to a succession of three convolutional layers. The first one convolves 32 8$\times$8 filters with stride four. The second layer convolves 64 4$\times$4 filters with stride two. The third hidden layer convolves 32 3$\times$3 filters with stride one. It ends with a fully connected layer with half as many output units as the chosen state representation dimension (because state representations have the form $\big[\varphi(\:o_{t-1}),\varphi(\:o_t)\big]$). The heads $\psi^k$ take as input the state representation and are composed of three fully connected layers, the two first ones of size $256$ and the last one of size two, which corresponds to the size of the action vectors (one torque per joint).

For SAC network architecture we choose a policy network that has the same structure as the heads $\psi^k$ used for imitation learning, also identical to the original SAC implementation~\citep{haarnoja2018soft}, and use the other default hyperparameters.

The Rectified linear units (ReLU) is used for the activation functions between hidden layers.
We use ADAM~\citep{kingma2014adam} with a learning rate of $10^{-4}$ to train the neural network $\varphi$, and $10^{-3}$ to train all the heads $\psi^k$ and the policy network.

\subsection{Results and Discussion}
\label{sec:res}

In this section, we report our results with the quantitative evaluation of SRLfD when a goal reaching task is used in the transfer learning phase (see \fig{fig:SRLfD}\subref{fig:SRLfD_c}).
Specifically, we evaluate the transferability of SRLfD learned state representations used as inputs for a RL algorithm (SAC~\citep{haarnoja2018soft}) to solve a new instance of the reaching task chosen randomly, and compare the success rates to the ones obtained with state representations originating from other methods.
The performance of a policy is measured as the probability to reach the goal from a random initial configuration in $50$ time steps or less. We expect that better representations yield faster learning progress and better convergence (on average).

\begin{figure}[t!]
\centerline{\includegraphics[trim = 0mm 0mm 0mm 0mm, clip, width=\textwidth]{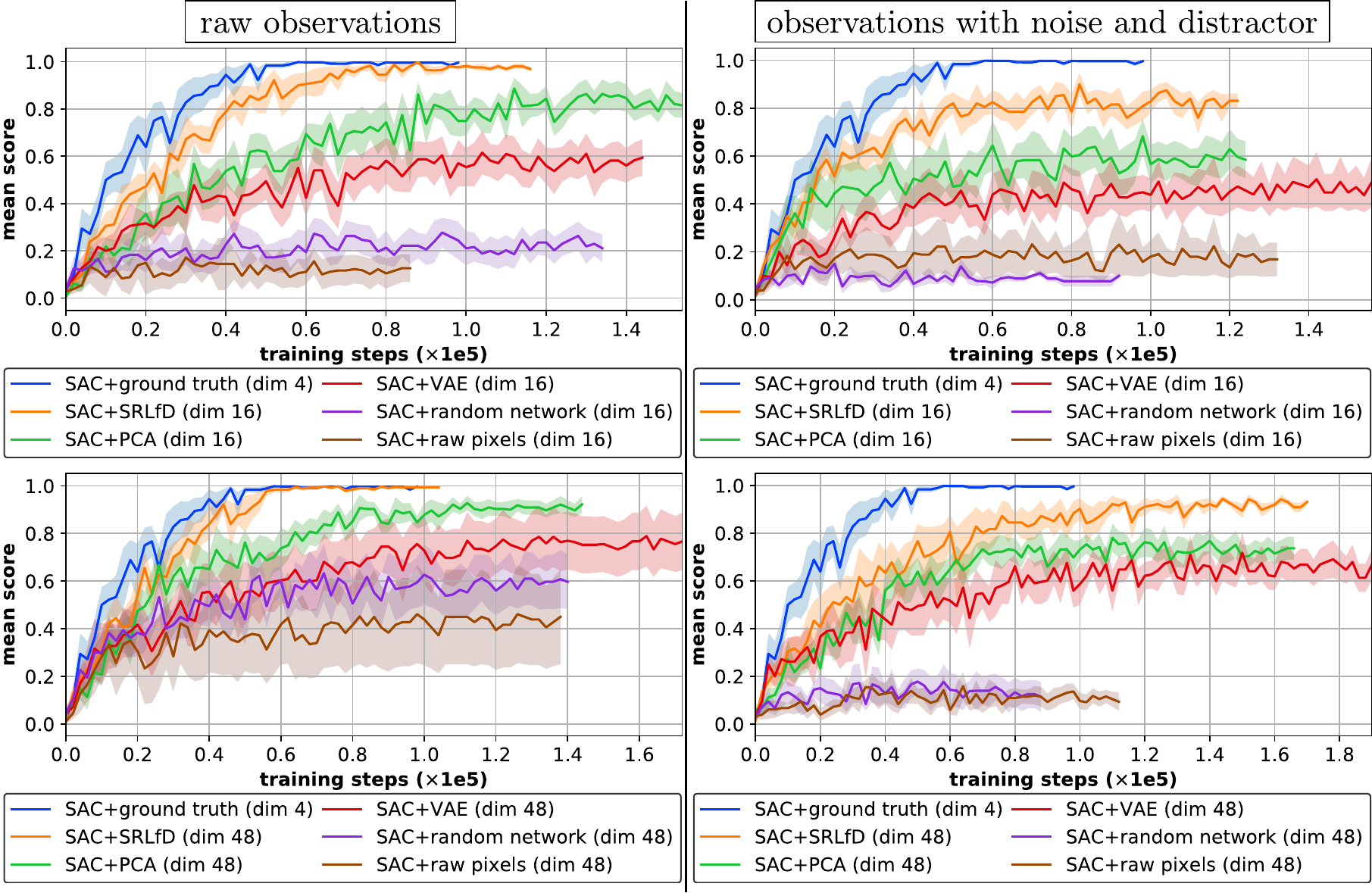}}
\caption{\label{fig:sac_baselinesall}
Learning curves of the episode returns averaged over 50 episodes (mean in lines and half standard deviation in shaded areas over 8 runs; the higher the better) with SAC algorithm with a random position of the goal for each run, based on various state representations. The indicated dimensions for SRLfD, PCA, VAE, random network and raw pixels correspond to the size of the state representation $\big[\varphi(\:o_{t-1}),\varphi(\:o_t)\big]$.
The use of our SRLfD state representation (red) in SAC outperforms all the other baselines, except the case in which SAC is given a direct access to the ground truth representation.}
\end{figure}

\begin{table}[t!]
\centering
\caption{Mean episode returns (mean $\pm$ standard deviation over 8 runs; the higher the better) corresponding to the end of the curves in \fig{fig:sac_baselinesall}.
\label{tab:RL_benchmark}}
\resizebox{0.8\columnwidth}{!}{%
\begin{tabular}{|c!{\vrule width 1.1pt}c|c|}
\hline
\textbf{Method} & \multicolumn{2}{c|}{\textbf{Mean score} } \\\cline{2-3}
 & \textbf{Raw observations} & \textbf{With noise and distractor} \\ \thickhline
\textbf{SAC+SRLfD (dim 48)}   & $\mathbf{0.992} \pm \mathbf{0.0080}$ & $\mathbf{0.928} \pm \mathbf{0.029}$ \\
\textbf{SAC+SRLfD (dim 16)}   & $\mathbf{0.980} \pm \mathbf{0.022}$ & $\mathbf{0.833} \pm \mathbf{0.082}$ \\ \hline
SAC+PCA (dim 48)  & $0.908 \pm 0.072$ & $0.725 \pm 0.10$  \\
SAC+PCA (dim 16)  & $0.832 \pm 0.13$ & $0.591 \pm 0.18$  \\ \hline
SAC+VAE (dim 48)  & $0.749 \pm 0.27$ & $0.650 \pm 0.13$  \\
SAC+VAE (dim 16)  & $0.574 \pm 0.16$ & {\color{RED}$0.448 \pm 0.18$}  \\ \hline
SAC+raw pixels (dim 48)   & {\color{RED}$0.365 \pm 0.39$} & {\color{RED}$0.106 \pm 0.063$}  \\
SAC+raw pixels (dim 16)   & {\color{RED}$0.118 \pm 0.13$} & {\color{RED}$0.149 \pm 0.14$}  \\\hline
SAC+random network (dim 48) & $0.552 \pm 0.21$ & {\color{RED}$0.143 \pm 0.14$}  \\
SAC+random network (dim 16) & {\color{RED}$0.239 \pm 0.13$} & {\color{RED}$0.0863 \pm 0.037$}  \\ \thickhline
SAC+ground truth (dim 4)  & $0.995 \pm 0.0054$ & $0.995 \pm 0.0054$ \\ \thickhline
\end{tabular}
}
\end{table}
\mytable{tab:RL_benchmark} displays the success rates corresponding to the end of the learning curves of \fig{fig:sac_baselinesall} obtained with SAC and different state representations in four different contexts: with a representation of either $16$ or $48$ dimensions (except for the ground truth representation, of size 4), and on ``clean'' \ie raw observations (in the middle of \fig{fig:reacher}) or observations with noise and a randomly moving ball distractor (on the right of \fig{fig:reacher}).
As expected, the best results are obtained with the ground truth representation, but we see that out of the five other state representations, only SRLfD, PCA, and VAE representations can be successfully used by SAC to solve reaching tasks when noise and a distractor are added to the inputs. SAC fails to train efficiently (in an end-to-end manner) the large neural network that takes raw pixels in input, whether its representation is of size 16 or 48. Using fixed random parameters for the first part of its network (random network representation) is not a viable option either.

The results show that with fewer dimensions our method (SRLfD) leads to better RL performances than with observation compression methods (PCA and VAE).
We assume that the information from the robotic arm can be filtered through the small size of the bottleneck due to the observation reconstruction objective (and dramatically more on the challenging observations).
This explains why PCA and VAE tend to require additional dimensions than the minimal number of dimensions of our robotic task (four dimensions: the two torques angles and velocities).
This clearly shows that with a carefully chosen unsupervised learning objective, such as the one used for SRLfD, it is possible to compact into a minimal number of dimensions only the information necessary for robotic control.

Another surprising observation is that PCA outperforms VAE in our results. By design, VAE is trained to encode and decode with as few errors as possible, and it can generally do this better than PCA by exploiting the nonlinearities of neural networks.
Moreover, as first explained by \citet{bourlard1988auto,kramer1991nonlinear}, the autoencoder is an extension of PCA that transforms correlated observations into nonlinearly uncorrelated representations. However, it is not clear that such uncorrelated input variables lead to better RL performances. This is because when data are obtained with transitions from a control system, the most important variables are those correlated with changes between transitions, which generally do not coincide with the directions of greatest variation in the data.

\section{Ballistic Projectile Tracking}
\label{sec:proj_tracking}

In this section, we study a transfer learning phase (see \fig{fig:SRLfD}\subref{fig:SRLfD_c}) corresponding to a simple supervised learning system to solve a ballistic projectile tracking task.
Specifically, it consists in training a tracker from learned representations to predict the next projectile position.
This task has the advantage of not needing $K$ oracle policies $\pi^k$ in a preliminary phase (\fig{fig:SRLfD}\subref{fig:SRLfD_a}).
Instead, we derive $\pi^k$ directly from the ballistic trajectory equations (\eq{eq:ballistic_trajectory}).
This enables us to easily perform the experimental study of the main hyperparameters of our SRLfD method: the state dimension $\mathcal{S}_d$ and the number of oracle policies $K$.
This also allows us to conduct a comparative quantitative evaluation against other representation strategies.
Furthermore, we study the possibility of using a recursive loop for the state update instead of the state concatenation. Thus, SRLfD can handle partial observability by aggregating information that may not be estimable from a single observation.
In particular, to solve a simple projectile tracking task, projectile velocity information is required and must be extracted from past measurements for the supervised learning algorithm to converge.
Mathematically, the observation (\ie measurement) $\:o_t$ is concatenated to the previous state estimate $\:s_{t-1}$ to form the input of SRLfD model $\varphi$ which estimates the current state as follows:
\begin{equation}
\:s_t \triangleq \varphi\left( \big[\:o_t, \:s_{t-1}\big] \right)
\end{equation}
where $\:s_{0}\sim\mathcal{N}\left(\:0_{\mathcal{S}_d},0.02\times\:I_{\mathcal{S}_d}\right)$.
Literally, this recursive loop conditions the current state estimate on all previous states.

An instance of this projectile tracking task is parameterized by the initial velocity and angle of the projectile.
Specifically, a tracker receives as input the state estimated by $\varphi$, and must predict the projectile's next position $\hat{\:o}_{t+1}$ as follows:
\begin{equation}\label{eq:tracker_pred}
\psi^{\text{new}}
\left(\varphi\left( \big[\:o_t, \:s_{t-1}\big] \right)\right)
= \hat{\:o}_{t+1}
\end{equation}
A tracker is then trained by supervised learning to minimize this objective function:
\begin{equation}
\mathcal{L} = \frac{1}{PT}\sum_{p=1}^P\sum_{t=1}^T
\Vert \psi^{\text{new}}\left(
\varphi\left( \big[\:o_t^{p}, \:s_{t-1}^{p}\big] \right)
\right) - \:o_{t+1}^{p} \Vert_2^2
\end{equation}
where the notations correspond to those defined in \mysec{sec:SRLfD_demonstrations}.

\subsection{Experimental Setup}

\paragraph{Baseline Methods}
We compare the state representations learned with SRLfD to five other representation strategies:
\begin{itemize}
    \item the ground truth is a vector of size 4 formed from the 2D cartesian positions and velocities of the projectile: $(x_t ,\, y_t ,\, v_{x,t},\, v_{y,t})$;
    \item the position corresponds to the 2D cartesian coordinates of the projectile: $(x_t ,\, y_t)$;
    \item a random network representation with the same $\varphi$ architecture and state recursive loop which is not trained\footnote{As in the previous reaching task, for random network representation the parameters of $\varphi$ are simply fixed to random values sampled from a Gaussian distribution of zero mean and standard deviation $0.02$.};
    \item an end-to-end representation learning strategy, \ie it builds its state estimate with the same $\varphi$ architecture and state recursive loop, while solving the tracking task;
    \item a Kalman filter estimated from positions with unknown initial cartesian velocities of the projectile.
\end{itemize}

The Kalman filter designed by \citet{kalman1960kalman} is a classical method for state estimation in state-linear control problems, where the ground truth state is not directly observable, but sensor measurements are observed instead.
\citet{kalman1960kalmanOC} create a mathematical framework for the control theory of the LQR (Linear-Quadratic-Regulator) problem and create for this purpose the notions of \emph{controllability} and \emph{observability} refined later in \citep{kalman1960general}.
\citet{witsenhausen1971separationControl} conducted one of the first attempts to survey the literature on the separation of state estimation and control.
In particular, this led to the two-step procedure, composed of the resolution of Kalman filtering and then of LQR, known as LQG (Linear-Quadratic-Gaussian).
However, the Kalman filter is also commonly used in recent RL applications \citep{ng2003autonomous,ng2006RLhelicopter,abbeel2007application,abbeel2008apprenticeship}.

The Kalman filter has then undergone many extensions including the popular extended Kalman filter (EKF) which can handle nonlinear transition models \citep{ljung1979asymptoticExKalman}.
However, a major drawback of these classical state estimation methods is that they require knowledge of the transition model.
This constraint has been relaxed by feature engineering (like SIFT \citep{lowe1999object} and SURF \citep{bay2006SURF}) techniques which require knowledge of the subsequent task \citep{kober2013surveyRL}.
Indeed, good hand-crafted features are task-specific and therefore costly in human expertise.
These drawbacks were then overcome by SRL methods popularized by \citet{jonschkowski2013learning}, which benefit from the autonomy of machine learning and the generalization power of deep learning techniques.

\paragraph{Kalman filter}
We briefly describe below the operations of the Kalman filter, the reader willing a complete presentation can refer to \citep{bertsekas2005dynamic}.
The equations for updating the positions ($x_{t},y_{t}$) and velocities ($v_{x,t},v_{y,t}$) of the projectile are related to their previous values, to the acceleration due to the gravitational force ($g$), and to the time elapsed between each update ($\Delta t$), as follows:
\begin{gather}
\begin{split}
& x_{t} = x_{t-1} + v_{x,t-1}\Delta t \\
& y_{t} = y_{t-1} + v_{y,t-1}\Delta t -\frac{1}{2}g(\Delta t)^2 \\
& v_{x,t}= v_{x,t-1}\\
& v_{y,t} = v_{y,t-1} - g\Delta t
\end{split}
\end{gather}
The ground truth state is defined as $\:s_t =\left[ x_t ,\, y_t ,\, v_{x,t},\, v_{y,t} \right] \in \mathbb{R}^4$.
Thus, the update procedure of the transition model is described with a single linear equation as follows:
\begin{align}\label{eq:dynamic_system}
\begin{split}
& \:s_t = \:A\:s_{t-1} + \:B\:u_{t-1} + \:w_{t-1} \\
& \:A = \begin{pmatrix}
1 & 0 & \Delta t & 0\\
0 & 1 & 0 & \Delta t \\
0 & 0 & 1 & 0 \\
0 & 0 & 0 & 1
\end{pmatrix} \quad , \qquad
\:B = \begin{pmatrix} 0 \\ \frac{(\Delta t)^2}{2} \\ 0\\ \Delta t \end{pmatrix} \quad , \qquad
\:w_{t-1}\sim\mathcal{N}\big(0,\:Q\big)
\end{split}
\end{align}
where $\:Q$ is the \emph{process noise covariance matrix} and $\:w_t$ is the \emph{process noise} assumed to be white (\ie normally, independently and identically distributed at each time step).

The observation of this transition model is the projectile position defined as
$\:o_t =\left[ x_t ,\, y_t \right] \in \mathbb{R}^2$ which is obtained from the ground truth state in the following way:
\begin{align}\label{eq:state2measure}
\begin{split}
\:o_t &= \:H \:s_t + \*\nu_t \quad,\qquad
\:H = \begin{pmatrix}
1 & 0 & 0 & 0 \\
0 & 1 & 0 & 0
\end{pmatrix} \quad,\qquad
\*\nu_t\sim\mathcal{N}\big(0,\:R\big)
\end{split}
\end{align}
where $\:R$ is the \emph{measurement noise covariance matrix} (\aka \emph{sensor noise covariance matrix}) and $\*\nu_t$ is the \emph{measurement noise} also assumed to be white. $\:R$ and $\:Q$ are ignored in our experiments for simplicity.

The aim of the Kalman filter is to solve the problem of estimating the ground truth state $\:s\in\mathbb{R}^{4}$.
To do this, the Kalman filter assumes to know $\:A$, $\:Q$, $\:R$ and $\:H$.
In this experiment, we assume there is no noise in the sensory inputs (\ie $\:R$ has only zero values), and no uncertainty in the transition model (\ie $\:Q$ has only zero values).
Moreover, in this projectile tracking task, there is no control vector ($a_t \triangleq 0$), because as the force exerted by gravitation is constant, it can be incorporated in the matrix $\:A$.
The state of the Kalman filter is initialized with the initial measurement (\ie the initial 2D cartesian position of the projectile) and zeros instead of the true initial velocities of the projectile.
In order to let the Kalman filter fix the initial velocities of the projectile and to ensure its convergence, we initialize the diagonal coordinates corresponding to the velocities of the \emph{state covariance matrix} (denoted $\:P$) to 100.

The Kalman filter follows a twofold procedure: (i) a prediction step which uses knowledge of the transition model, (ii) an update step which combines the model and the measurement knowing that both may be imperfect.
The equations of the prediction step consists in the prediction of the state estimate \eq{eq:kalman_s_pred}, and the prediction of the state covariance estimate \eq{eq:kalman_P_pred}.
\begin{align}
& \hat{\:s}_t^{-} = \:A \hat{\:s}_{t-1} + \:B\:u_{t-1}\label{eq:kalman_s_pred} \\
& \:P_t^{-} = \:A \:P_{t-1} \:A^{\tau} + \:Q\label{eq:kalman_P_pred}
\end{align}
This corresponds to a priori estimates.

The equations in the update step first update the Kalman gain matrix \eq{eq:kalman_K}, then update the state estimate by incorporating the measurement and the a priori state estimate \eq{eq:kalman_s_update}, and similarly update the state covariance matrix \eq{eq:kalman_P_update}.
\begin{align}
& \:K_t = \:P_t^{-} \:H^{\tau} \big[\:H \:P_t^{-} \:H^{\tau} + \:R \big]^{-1} \label{eq:kalman_K}\\
& \hat{\:s}_t = \hat{\:s}_t^{-} + \:K_t \big(\:o_t - \:H \hat{\:s}_t^{-} \big) \label{eq:kalman_s_update}\\
& \:P_t = \big(\:{Id} -\:K_t \:H \big)\:P_t^{-}\label{eq:kalman_P_update}
\end{align}
Therefore, at each iteration, the Kalman filter predicts the next a priori state estimate which is then used to update the next state estimate which corresponds to an a posteriori estimate.
This implies that the Kalman filter is a recursive linear state estimation, whereas SRLfD is a recursive nonlinear state estimation (thanks to the state recursive loop).
In particular, while SRLfD may use a linear network $\varphi$ for simple tasks, it may still take advantage of nonlinear approximations such as multilayer perceptrons defining its $\psi^k$ heads, such as in this ballistic projectile tracking task.

\paragraph{Generating Demonstrations}
\begin{figure}[t!]
\centerline{\includegraphics[width=0.7\columnwidth]{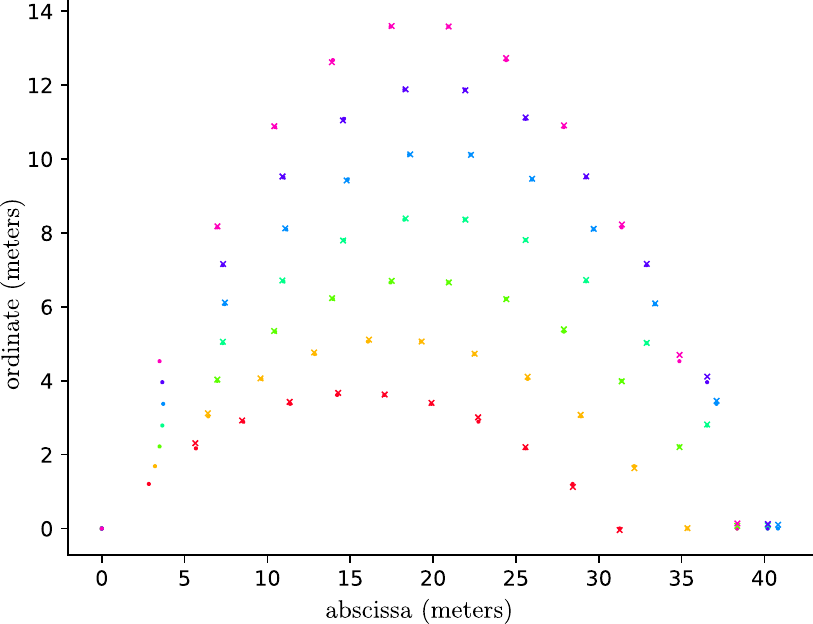}}
\caption{\label{fig:eval_ballistic}
Predictions (crosses) with a tracker trained from SRLfD representations, over 7 trajectories with zero initial coordinate (\ie $x_0=0$ and $y_0=0$), initial angles $\alpha_0\in\{25,30,35,40,45,50,55\}$ (in degrees), and initial velocity 20 (in meter per second), composed of 12 projectile position measurements (points).
}
\end{figure}
The projectile tracking task does not require in its preliminary phase $K$ oracle policies $\pi^k$.
It is the ballistic trajectory equations (\eq{eq:ballistic_trajectory}) that allow us to define these oracle policies $\pi^k$.
As we ignore all forces except the gravitational one, the trajectory of the projectile corresponds to a ballistic trajectory.
The temporal equations of the ballistic trajectory are defined with the force of gravity $g=9.81$ m/s$^2$, the initial angle of launch of the projectile $\alpha_0$, and its initial velocity $v_0$ as well as its initial y-coordinate $y_0$, as follows:
\begin{align}\label{eq:ballistic_trajectory}
\begin{split}
& x\left(t\right)= v_0\cos\left(\alpha_0 \right)t \\
& y\left(t\right)=-{\frac {1}{2}}g t^{2}+ v_0 \sin\left(\alpha_0 \right) t+ y_0 \\
& v_x\left(t\right)= v_0 \cos \left(\alpha_0 \right)\\
& v_y\left(t\right)= -gt + v_0 \sin\left(\alpha_0 \right)
\end{split}
\end{align}
The total horizontal distance $x_{\f{max}}$ covered until the projectile falls back to the ground is given by:
\begin{equation}\label{eq:ballistic_max_x}
x_{\f{max}}={\frac {v_0}{g}}\cos \alpha_0 \left(v_0\sin \alpha_0 +
{\sqrt {(v_0\sin \alpha_0 )^{2}+2g y_0}}\right)
\end{equation}
This allows us to calculate the corresponding time of flight $t_{\f{max}}$:
\begin{equation}\label{eq:ballistic_max_t}
t_{\f{max}}=\frac{x_{\f{max}}}{v_0\cos \alpha_0}
\end{equation}
These equations provide an oracle policy parameterized by $v_0$ and $\alpha_0$, which can generate ballistic trajectories at any initial ordinate $y_0$.
Each generated trajectory is of fixed length $T=12$ which corresponds to 10 demonstration samples (as the first two are used during initialization) of the form $(\:o_t^{k,p}, \:a_t^{k,p})$ where the actions $\:a_t^{k,p}$ correspond to the next positions of the projectile, \ie $\:a_t^{k,p} = \:o_{t+1}^{k,p}$.
To do this, we define for each trajectory the time between each update $\Delta t$, as follows:
\begin{equation}\label{eq:delta_t}
\Delta t = \frac{t_{\f{max}}}{T-1}
\end{equation}
where $t_{\f{max}}$ is defined in \eq{eq:ballistic_max_t}.

For the pretraining phase of SRLfD (see \fig{fig:SRLfD}\subref{fig:SRLfD_b}), each oracle policy $\pi^k$ has a fixed initial velocity and angle ($v_0^k, \alpha_0^k$), and generate $P$ ballistic trajectories with different random initial ordinates such that $y_0\in[0,\, 30]$.
We uniformly pick $K'$ tasks (such that $K'\leq K$) to simultaneously train the corresponding heads $\psi^k$ for computational efficiency.
Each of them is trained on demonstrations sequentially generated from $P=\frac{B}{K'}$ different paths where $B$ is a desired batch size for training $\varphi$.
This way, every optimization iteration to train $\varphi$ is performed on a fixed number $B$ of demonstrations, independently of the total number of tasks $K$.
We use $B=256$ and $K'=\f{min}(K,6)$ in our experiments.
For the tracker training, the batch size is also 256, which implies $P=256$.

For the SRLfD training validation, we measure the average of tracking prediction errors over all oracle policies on initial ordinates defined as $y_0 \in \{0,10,20,30\}$ (in meters).
For the tracker training validation, we measure the average of tracking prediction errors on fixed trajectories with the same initial velocity $v_0=20$ and the initial angles defined
as $\alpha_0\in\{25,30,35,40,45,50,55\}$ (in degrees) and on initial ordinates defined as $y_0 \in \{0,10,20,30\}$ (in meters).
\fig{fig:eval_ballistic} shows some qualitative results of this validation with a tracker trained from 4-dimensional SRLfD representations with 6 oracle policies.

\paragraph{Implementation Details}

$\varphi$ is a linear neural network of input dimension $(2~+~\mathcal{S}_d)$ and output dimension $\mathcal{S}_d$.
When not specified, the default number of oracle policies $K$ is 6.
We used a state recursive loop to remove the state concatenation.
For the random network and the end-to-end baselines, the networks have the same structure as for $\varphi$, where in the former the parameters are kept fixed, while in the latter $\varphi$ is trained jointly with the tracker $\psi^{\text{new}}$.
The heads $\psi^k$ and the tracker $\psi^{\text{new}}$ are one-hidden neural networks, with the hidden one of size $32$ and the last one of size two, which corresponds to the size of the action vectors (\ie the next projectile positions).
The previous nonlinear networks are necessary because $\Delta t$ changes on all ballistic trajectories, so unlike the Kalman filter which knows this value, for the tracker and SRLfD heads they must relate the input to the output nonlinearly.

We use ADAM~\citep{kingma2014adam} with a learning rate of $10^{-4}$ to train $\varphi$, the heads $\psi^k$, and the tracker $\psi^{\text{new}}$.
For the SRLfD and tracker trainings, we use early stopping of patience 40 epochs.
In one epoch 10 000 iterations are performed, $\varphi$ (during SRLfD training) or $\psi^{\text{new}}$ (during tracker training) see exactly 1 000$\times$256 different trajectories composed of 10 demonstrations (\ie data points or samples).
The Leaky Rectified Linear Unit (Leaky ReLU) is used for the activation function \citep{xu2015activations}.

\subsection{Results and Discussion}

\begin{figure}[t!]
\centering
\subcaptionbox{
\label{fig:boxplot_bestMethods_eval}}{\includegraphics[width=0.4\linewidth]{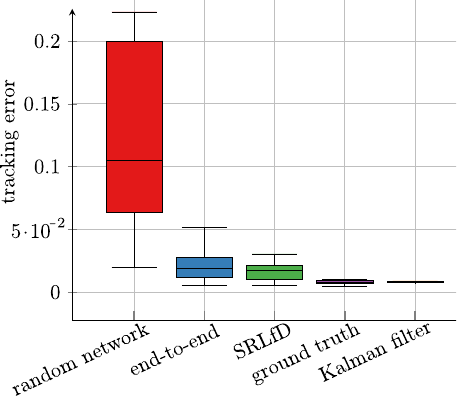}}
\subcaptionbox{
\label{fig:boxplot_SRLfD_nPi_eval}}{\includegraphics[width=0.4\linewidth]{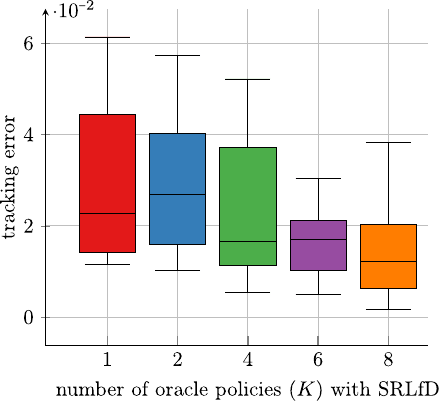}}
\\
\subcaptionbox{ \label{fig:boxplot_SRLfD_stateDim_eval}}{\includegraphics[width=0.4\linewidth]{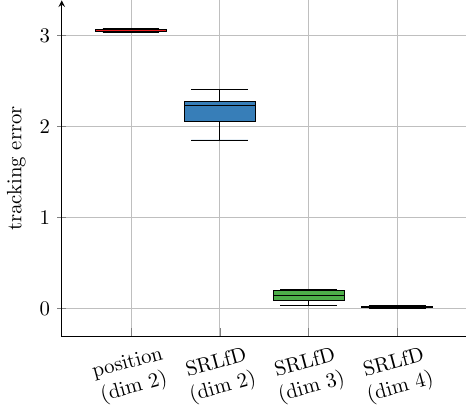}}
\subcaptionbox{ \label{fig:boxplot_SRLfD_stateDim_eval_all}}{\includegraphics[width=0.4\linewidth]{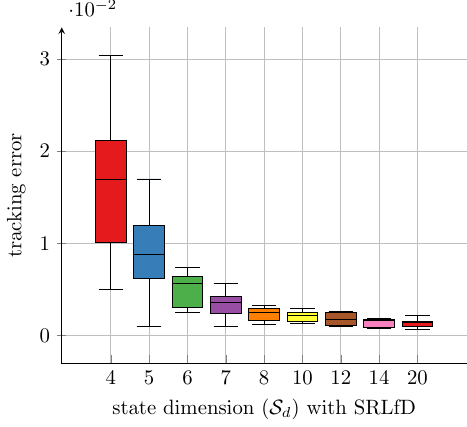}}\hspace{0.cm}
\caption{\label{fig:all_boxplots}
Mean tracking errors over last 5 epochs (average on 10 runs; the lower the better) obtained from five representation strategies with different hyperparameters for SRLfD.
\textbf{\subref{fig:boxplot_bestMethods_eval}} Five representation strategies of size 4.
\textbf{\subref{fig:boxplot_SRLfD_nPi_eval}} SRLfD representations of size 4 with different number of oracle policies $K$.
\textbf{\subref{fig:boxplot_SRLfD_stateDim_eval}} Representation strategies of size 4 and lower.
\textbf{\subref{fig:boxplot_SRLfD_stateDim_eval_all}} SRLfD representations with 6 oracle policies of varying sizes $\mathcal{S}_d$.}
\end{figure}

We learned the projectile tracking task from six representation strategies.
\fig{fig:all_boxplots}\subref{fig:boxplot_bestMethods_eval} shows the boxplot of the average tracking errors obtained with all different strategies of the same size 4.
Our SRLfD outperforms the end-to-end representation, confirming the empowerment provided by ``divide and conquer'' techniques \citep{dasgupta2008algorithms}.
Ground truth representations outperform our method because they know the complete projectile configuration.
With a random network, the supervised learning system fails to track the projectile, implying that state representation learning is required in the context of recursive state estimation.

Regarding the Kalman filter, it uses the knowledge of the real transition model in order to provide a complete recursive state estimation.
It is therefore not surprising that it achieves the same performance as the ground truth baseline.
Unlike this classical method, SRLfD does not use the a priori knowledge of the tracking task but only that available in the oracle policies.
The performance obtained with the SRLfD representations in this comparative evaluation shows that they extract the position and velocity of the projectile.
In other words, there is enough diversity in the oracle policies to be imitated by SRLfD network heads, so that their joint state space is close to the real state that makes the system fully observable.
The comparative quantitative evaluation presented by \fig{fig:all_boxplots}\subref{fig:boxplot_SRLfD_nPi_eval} confirms this hypothesis since the performance obtained with the size 4 SRLfD representations increases with the number of oracle policies used during the pretraining phase of SRLfD.

\mytable{tab:tracking_tables} displays the average tracking errors displayed in \fig{fig:all_boxplots}, with the mean and standard deviation for better insight.
\fig{fig:all_boxplots}\subref{fig:boxplot_SRLfD_stateDim_eval} shows that as the state dimension decreases, the information lost by SRLfD significantly degrades the performance of the trained trackers, while for a size of 2, it is still better than the position baseline.
On the other hand, \fig{fig:all_boxplots}\subref{fig:boxplot_SRLfD_stateDim_eval_all}
shows that as the state dimension increases, the performance of the trained trackers improves until it even outperforms ground truth starting at 6 dimensions (see \mytable{tab:tracking_tables}).
Although these results may be surprising, one can assume that adding redundancy to the representations makes them easier to build (since the dimension of the ground truth vector is 4).
Indeed, larger state embeddings could be more regular and thus be build with simpler neural networks which are less subject to the overfitting problem.
However, the question of what is an ideal representation for deep learning algorithms is far from being answered.
Recently works have started to investigate this question \citep{ota2020can,ota2021trainingDRL}, but the search for a definitive answer leads far beyond the scope of this work.

\begin{table}[!htb]
\caption{\label{tab:tracking_tables}
Tracking errors corresponding to \fig{fig:all_boxplots} (mean $\pm$ standard deviation on 10 runs; the lower the better):
\textbf{\subref{tab:trackingErrors_baselines}} obtained with the five baselines,
\textbf{\subref{tab:trackingErrors_policies}} obtained with SRLfD representations of size 4 with different number of oracle policies $K$,
\textbf{\subref{tab:trackingErrors_dims}} obtained with SRLfD representations with 6 oracle policies of varying sizes $\mathcal{S}_d$.
}
\begin{subtable}[]{1\linewidth}
\centering
\caption{\label{tab:trackingErrors_baselines}}
\begin{tabular}{|c!{\vrule width 1.1pt}c|}
\hline
\textbf{Method} & \textbf{Tracking error ($\times 10^{-3}$)} \\ \hline
Ground truth (dim 4) & $7.97 \pm 1.64$ \\ 
Kalman filter (dim 4) & $8.14 \pm 0.6$ \\
Position (dim 2)  & $3 \, 000 \pm 26.9 $ \\
End-to-end (dim 4) & $ 22.4 \pm 13.7$ \\
Random network (dim 4) & $ 130 \pm 74.5$ \\ \hline
\end{tabular}
\end{subtable}%
\\
\begin{subtable}[t]{.5\linewidth}
\caption{\label{tab:trackingErrors_policies}}
\centering
\begin{tabular}{|c!{\vrule width 1.1pt}c|}
\hline
$K$ & \textbf{Tracking error} ($\times 10^{-3}$)\\ \hline
1 & $31.5 \pm 19.7$  \\
2 & $31.4 \pm 16.1$  \\
4 & $26.3 \pm 16.4$  \\
6 & $\mathbf{17.1} \pm 7.46$  \\
8 & $18.4 \pm 16.8$  \\ \hline
\end{tabular}
\end{subtable}%
\hfill 
\begin{subtable}[t]{.5\linewidth}
\caption{\label{tab:trackingErrors_dims}}
\centering
\begin{tabular}{|c!{\vrule width 1.1pt}c|}
\hline
$\mathcal{S}_d$ & \textbf{Tracking error ($\times 10^{-3}$)} \\ \hline
2 & $2 \, 200 \pm 175 $  \\
3 & $147 \pm 60.4$  \\
4 & $17.1 \pm 7.46$  \\
5 & $9.37 \pm 4.56$  \\
6 & $5.21 \pm 1.78$  \\
7 & $3.61 \pm 1.37$  \\
8 & $2.41 \pm 0.733$  \\
10 & $2.16 \pm 0.574$  \\
12 & $1.83 \pm 0.686$  \\
14 & $1.44 \pm 0.44$  \\
20 & $\mathbf{1.38} \pm 0.432$  \\ \hline
\end{tabular}
\end{subtable}%
\end{table}

\section{Conclusion}
\label{sec:ccl}

We presented a method (SRLfD) for learning state representations from demonstrations, more specifically from runs of oracle policies on different instances of a task. Our results indicate that the learned state representations can advantageously replace raw sensory inputs to learn policies on new task instances via regular RL. By simultaneously learning an end-to-end technique for several tasks sharing common useful knowledge, SRLfD forces the state representation to be general, provided that the tasks are diverse. Moreover, since the representation is trained together with heads that imitate the oracle policies, we believe that it is more appropriate for control than other types of representations (for instance ones that primarily aim at enabling a good reconstruction of the raw inputs).
Our experimental results tend to confirm this belief, as SRLfD state representations were exploited more effectively by the SAC RL algorithm and a supervised learning system, than several other types of state representations.

\section*{Acknowledgments}
This article has been supported within the Labex SMART supported by French state funds managed by the ANR within the Investissements d'Avenir program under references ANR-11-LABX-65 and ANR-18-CE33-0005 HUSKI.
We gratefully acknowledge the support of NVIDIA Corporation with the donation of the Titan Xp GPU used for this research.

%
%
%
\bibliographystyle{apalike} 
\bibliography{references.bib}  
\end{document}